\documentclass[letterpaper, 10 pt, conference]{ieeeconf}  

\pdfoutput=1

\IEEEoverridecommandlockouts                              

\usepackage{amsmath, amsthm, amsfonts, amssymb,color}
\usepackage[ruled,vlined,noend]{algorithm2e}
\usepackage{blindtext}
\usepackage{graphicx}
\usepackage[utf8]{inputenc}
\usepackage{mathtools}
\usepackage{tikz, cite}
\usepackage[font=footnotesize,labelformat=simple]{subcaption}
\usepackage{tabu}

\newtheorem{remark}{Remark}
\newtheorem{definition}{Definition}
\newtheorem{example}{Example}
\usepackage{epsfig} 
\usepackage{psfrag}
\usepackage{mathrsfs}

\usepackage{multirow}
\usepackage{array,booktabs,arydshln,xcolor}
\usepackage{soul}
\usepackage{mathtools}
\usepackage{nccmath}
\usepackage{caption}
\usepackage{nccmath}
\usepackage{comment}
\usepackage{pdfpages}
\usepackage{comment}
\renewcommand{\vec}[1]{\boldsymbol{#1}}
\usepackage{float}
\usepackage{graphicx}
\usepackage{epstopdf}
\usepackage{color}
\usepackage{verbatim}
\usepackage{tikz,times}
\usepackage{xcolor}
\usepackage{lipsum}
\usepackage{makecell}
\let\labelindent\relax

\usepackage{enumitem}
\usepackage{multirow}

\newtheorem{assumption}{\textbf{Assumption}}

\usepackage{stfloats}

\title{\LARGE \bf
Motion Planning under Uncertainty: Integrating Learning-Based Multi-Modal Predictors into Branch Model Predictive Control
}

\author{Mohamed-Khalil Bouzidi$^{1,3}$, Bojan Derajic $^{2,3}$,  Daniel Goehring$^{1}$, Joerg Reichardt$^{3}$
\thanks{$^{1}$  Free Universitity of Berlin, Germany }
\thanks{{\tt\small \{firstname.lastname@fu-berlin.de\}}}
\thanks{$^{2}$ Technical Universitity of Berlin, Germany }
\thanks{{\tt\small \{firstname.lastname@tu-berlin.de\}}}
\thanks{$^{3}$ Continental AG}
\thanks{{\tt\small \{firstname.lastname@continental.com\}}}
\thanks{This work is funded by the German Federal Ministry for}
\thanks{Economic Affairs and Climate Action within the project "nxtAIM".}
}

\begin{document}

\maketitle
\thispagestyle{empty}
\pagestyle{empty}

\begin{abstract}
In complex traffic environments, autonomous vehicles face multi-modal uncertainty about other agents' future behavior. To address this, recent advancements in learning-based motion predictors output multi-modal predictions.  We present our novel framework that leverages Branch Model Predictive Control(BMPC) to account for these predictions. The framework includes an online scenario-selection process guided by topology and collision risk criteria. This efficiently selects a minimal set of predictions, rendering the BMPC real-time capable. Additionally, we introduce an adaptive decision postponing strategy that delays the planner's commitment to a single scenario until the uncertainty is resolved. Our comprehensive evaluations in traffic intersection and random highway merging scenarios demonstrate enhanced comfort and safety through our method.
\end{abstract}
\section{Introduction} \label{sec:Intro}

\begin{figure}[t]
\centering
    \includegraphics[width=0.49\textwidth]{"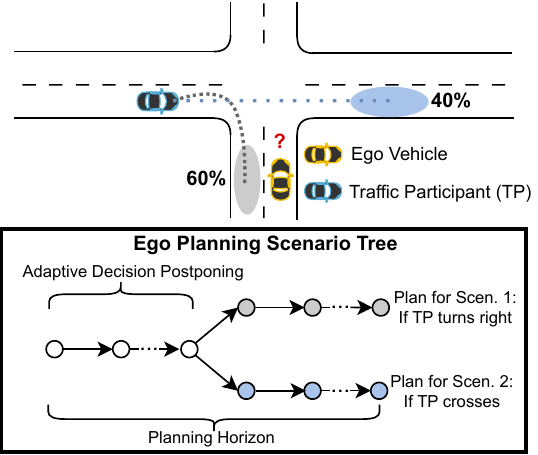"}
    \caption{An illustration of the challenge of planning in  traffic scenes with uncertain multi-modal predictions. To address this, Branch MPC optimizes over a scenario tree.}
    \label{fig:P1}
\vspace{-1.5em}    
\end{figure}

Autonomous vehicles (AVs) must anticipate and predict the behavior of other traffic participants (TPs) to plan safe and comfortable trajectories.  In dynamic environments, however, predictions of how a traffic scene will evolve are quite uncertain over the entire planning horizon. This uncertainty arises  from the unknown TP intentions  but also from other factors such as  sensor noise and occlusion. In addition, the probability distributions for the development of a traffic situation are often multi-modal, i.e., at the end of the planning horizon, several quite different environmental states may be similarly probable (s. Fig \ref{fig:P1}). Consequently, planning algorithms must find a plan that may not be optimal for any single future scenario, but rather represents a compromise against all possibilities of the future.

Recently, various approaches are proposed to address such cases.
In \cite{hubmann_automated_2018,  somani_despot_2013} uncertainty-aware planners are presented where the problem is formulated as a Partially  Observable Markov Decision Process (POMDP).  These approaches suffer from the fact that solving POMPD with increasing time horizon and action space is no longer real-time capable. For this reason, the action space and time steps are only coarsely 
sampled, causing suboptimal solutions hindering their suitability for dealing with highly dynamic environments. 
On the other hand, there are some variations of Model Predictive Control (MPC), which also take into account the 
uncertainty.
One approach is robust MPC \cite{schildbach_collision_2016, singh}. These methods consider all possible cases or the worst case to  ensures collision avoidance in all possible scenarios. However, these approaches lead to conservative, overcautious behaviors, possibly preventing the AV from moving forward at all. To reduce conservatism, stochastic MPCs  \cite{benciolini_multistage_2021, brudigam_combining_2018} (or 
variants such as scenario-based MPC \cite{kensbock_scenario-based_2023,de_groot_scenario-based_2021}) are developed to exclude very unlikely predictions and guarantee constraints satisfaction with a certain probability.
Still, this can be very conservative, especially with multi-modal uncertainty. The main source of conservatism is that the planner ignores future information gathering about the scene that the AV can use, i.e., it does not take into account feedback in its predictions. 
For instance, in the scene depicted in Fig.\ \ref{fig:P1} an experienced driver would be aware that in some timesteps, he will get more information about which mode will occur (i.e., whether the TP will turn right or cross the street) and use it for his motion plan. 
 In contrast, the mentioned robust or stochastic MPC plan to comply with both modes for the entire planning horizon. This concept of future feedback is included in the so-called Branch 
MPC which can plan with different predictions of the traffic by allowing different control inputs/strategies for different scenario evolutions. 

Branch MPC for Motion planning and Control task has already been implemented in \cite{oliveira_interaction_2023, chen_interactive_2021, alsterda_contingency_2019, fors_resilient_2022,  ulfsjoo_integrating_2022,  tas_decision-theoretic_2023}.
However, the integration of TP's predictions  in these works were restricted to rule-based predictors\cite{oliveira_interaction_2023, chen_interactive_2021} or model-based predictors \cite{ fors_resilient_2022, ulfsjoo_integrating_2022}. These prediction methods lack accuracy and flexibility due to rigidity or incompleteness of the underlying model or rulesets (e.g., the proposed rule-based approaches can only take one TP into account), limiting the prediction quality in scenes with complex interaction of heterogeneous agents.   In contrast, learning-based methods excel due to their ability to handle complex and uncertain nature of other TP's behavior extracting patterns from high-dimensional data and providing multi-modal predictions. Consequently, recent works are predominantly focused on learning-based predictors \cite{shi_motion_2022, nayakanti_wayformer_2022, qcnet}. 
However, integrating these predictors poses a challenge for which viable solutions have yet to be proposed. Branch MPC struggles with scalability regarding the number of scenarios which does not allow to use the raw output of these predictor. Therefore, we propose a novel approach to select only relevant predictions for the scenario tree based on topology concepts and collision risk. Additionally, we further leverage the learning-based output to design a decision postponing strategy by adaptively setting the branching time (s. fig \ref{fig:P1}). This parameter significantly impacts planner performance but is typically treated as a tuning parameter selected offline in previous works.
The main contributions of our work are therefore summarized as:
\begin{itemize}[leftmargin=*]
    \item  An uncertainty-aware Motion Planning framework integrating learning-based predictor into Branch MPC
    \item A scenario selection strategy based on topology concepts and collision risk 
    \item An adaptive decision postponing strategy to adapt the branching time based on the predictions

\end{itemize}
The remainder of the paper is organized as follows: We introduce the employed multi-modal Predictor and the Baseline MPC in Section II. In Section III, we augment the Baseline MPC to the Branch MPC Framework.  Section IV presents our strategy to select and integrate the relevant predictions of the TPs into the Branch MPC. Section V presents our adaptive decision postponing method. The evaluation is provided in Section V. Section VI concludes the paper.

\section{Preliminaries} \label{sec:pre}
Consider a motion planning problem of an AV in an arbitrary traffic scene with $O$ TPs $o \in \{0,..,O-1\}$ (such as pedestrians, cars, bicycles). The AV lacks knowledge about the intentions and future trajectory of TPs.  It relies solely on current and past states obtained through its environment perception and tracking module. Additionally, a map $M_{in}$ containing polylines with the respective road attributes of each point (such as location and lane type) is given. A reference path with road boundaries is then extracted from $M_{in}$ possibly by a high-level route planner. This reference path denoted as $\mathcal{P}_{ref}: [0,\theta_{max}] \to \mathbb{R}^4 \times [0, 2\pi]$ maps the arclength $\theta$  to a tuple consisting of the reference pose and distances to the left and right road boundaries $\theta \mapsto (x_{ref}(\theta), y_{ref}(\theta), d_{lb}(\theta), d_{rb}(\theta), \psi_{ref}(\theta))$ where $\theta_{max}$ is the maximum arclength.

\subsection{Learning-based Multi-Modal Motion Predictor} \label{sec:predictor}
To effectively operate in a scene with TPs, the AV must anticipate their behavior using a prediction module. 
 Due to the TP's unknown intentions and range of potential choices, this needs to be inferred probabilistically.
Consequently, most SotA learning-based predictors provide multi-modal distributed predictions for different possible behaviors of TPs. 
Catalyzed by the availability of several large-scale datasets of traffic scenarios, a wide range of these approaches has been developed in the recent literature.
One of these predictors is Motion Transformer (MTR) \cite{shi_motion_2022}.
We employ the MTR, noting that any other predictor capable of multi-modal predictions could serve as an alternative. The MTR is built using a transformer-based context encoder and motion decoder.
The inputs to the model are the history of position, heading angle, and velocity of all surrounding TPs over a specified time frame. Additionally, it incorporates the road map $M_{in}$ including the lane segment attributes to provide scene context. Based on these inputs, the MTR generates multiple predictions regarding how TPs could potentially interact.
The MTR's output is structured as a tuple of Gaussian Mixture Models (GMMs) for each object's future state $\mathcal{N}^o_{k,m}(\vec{\mu}^o_{k,m}, \boldsymbol{\Sigma}^o_{k,m},\pi^o_m)$ at each time step within the prediction horizon $N$. Each component $m \in \mathcal{M}$ of this mixture represents a predicted mode. Hence, the predicted distribution of each TP for each timestep $k$ can be expressed as:
\begin{equation}\label{eq:gmm}
p^o_{k} = \sum_{m \in \mathcal{M}}  \pi^o_m \mathcal{N}^o_{k,m}(\vec{\mu}^o_{k,m}, \boldsymbol{\Sigma}^o_{k,m})
\end{equation}
Conversely, the predicted trajectories of each object $o$ are extracted from the GMM along with the respective probability $\pi^o_m$ of each mode $m$. These consist of the predicted states for each timestep $k$, denoted as $\vec{o}_{k,m} = [x^o_{k,m}, y^o_{k,m}, \psi^o_{k,m}, v^o_{i,m} ]^\top$,  including position, heading, and velocity.

\subsection{Nominal Model Predictive Contouring Control}\label{sec:nominal}
Model Predictive Contouring Control (MPCC) as introduced in \cite{brito1}  does not take into account multi-modal predictions of the TPs and only plans with one (usually with the highest probability $\pi^o_m$) prediction. As it serves as our baseline that we will extend into Branch MPCC in the next section, we now briefly recall its approach.

For the state space model $\dot{\vec{z}}(t) = f(\vec{z}(t), \vec{u}(t))$ we use the kinematic bicycle model:
\begin{equation}\label{eq:2}
    \dot{\vec{z}} = \left[ v \cos(\psi), \ v \sin(\psi), \ v \frac{\tan(\psi)}{l}, \ a, \ j, \ \dot{\delta}, \dot{\theta}  \right]^\top
\end{equation}
where $\vec{z} = [x,y,\psi, v, a, \delta, \theta ]^\top$ is the state vector and $\vec{u} = [j, \dot{\delta},\dot{\theta}]^\top $  the control input vector. The acceleration, steering angle, jerk, steering angle rate, virtual speed projected on the path, and wheelbase are denoted as $ a, \delta, j, \dot{\delta},\dot{\theta}, l $, respectively.
Variables with a subscript $k$ denote the timestep $k$ of the discretized variables defined in eq. (\ref{eq:2}).

The MPCC maximizes path progress while  minimizing lag error $\hat{e}^l_k$ and contouring error $\hat{e}^c_k$, approximated by:

\begin{equation}
\begin{aligned}
& \left[\begin{array}{cr}
\hat{e}^c_k \\
\hat{e}^l_k 
\end{array}\right]
=\left[\begin{array}{cr}
\sin (\psi_{ref} \left(\theta_k\right)) & -\cos (\psi_{ref} \left(\theta_k\right)) \\
-\cos (\psi_{ref} \left(\theta_k\right)) & -\sin  (\psi_{ref} \left(\theta_k\right))
\end{array}\right]
\Delta \vec{p}_{ref}\nonumber
\end{aligned}
\end{equation}
where $\Delta \vec{p}_{ref} = [x-x_{ref}\left(\theta_k\right), \ y-y_{ref}\left(\theta_k\right)]^\top $.
This allows us to define the running cost:
\begin{equation}\label{eqn:J}
    J_{k} = \left[\begin{array}{cr}
\hat{e}^c_k \\
\hat{e}^l_k 
\end{array}\right]^\top Q 
\left[\begin{array}{cr}
\hat{e}^c_k \\
\hat{e}^l_k 
\end{array}\right]
-q_v v_k^p + \vec{u}_k^\top R \vec{u}_k + J^p_{k}
\end{equation}
where $Q, q_v, R$ are the respective weights. 
Additionally, we include the cost term $J^p_{k}$, employing the Potential Field method \cite{siebenrock}, to address the uni-modal uncertainty (i.e. of one mode) inherent in the predicted TP trajectories. 

\begin{equation}\label{eqn5_7}
    \begin{split}
    \centering
    J^p_{k} &= q_{ob} \sum_{i=0}^{O-1} \cdot \exp\left(-\left(\frac{\Delta x^o_k}{l^o}\right)^2 - \left(\frac{\Delta y^o_k}{w^o}\right)^2\right) \\
    & + q_{lm} \sum_{l=0}^{L-1} \exp\left(-\left(\frac{d_{lm}^l-\hat{e}^c_k\left(\theta_k\right)}{\sigma}\right)^2\right)
    \end{split} 
\end{equation}
where $\Delta x^i_k, \Delta y^i_k$ are the distances to the respective obstacle, $d_{lm}^l$ is the signed distance from the reference path to the respective $L$ lane marker, $\sigma$ a scaling factor and $l^i, w^i$ a conservative estimation of the length and width of the obstacle considering the uni-modal uncertainty and $q_{ob}, q_{lm}$ are the respective weights. 
This supplements the existing hard constraints for obstacle avoidance and road boundaries.

For feasible vehicle trajectories, we add the actuator constraints $j_k \in [j_{\text{min}}, j_{\text{max}}]$, $\dot{\delta}_k \in [\dot{\delta}_{\text{min}}, \dot{\delta}_{\text{max}}]$ and $\delta_k \in [\delta_{\text{min}}, \delta_{\text{max}}]$ and limit longitudinal and lateral acceleration based on \cite{polack} .



\begin{figure*}[t]
\vspace{0.6em} 
    \centering
    \includegraphics[width=1\textwidth]{"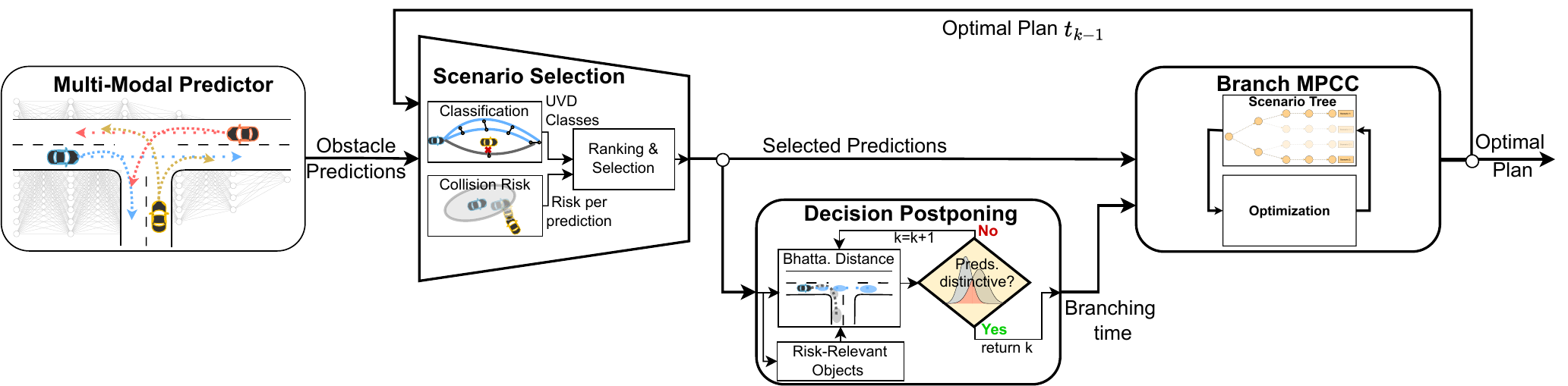"}
    \caption{Our learning-aided Motion Planning and Control Framework  using a Multi-modal Predictor, Scenario Selection, Adaptive Decision Postponing and Branch MPCC}
    \label{fig:method}
    \vspace{-1.2em} 
\end{figure*}
\section{Branch Model Predictive Contouring Control} \label{sec:BMPCC}
To account for multi-modal predictions, we augment the MPCC. 
 
The Branch MPCC (BMPCC) constructs a scenario tree to represent the uncertainty in the scene.  Each branch of the tree corresponds to a distinct prediction (mode) of the traffic scene evolution, i.e., we consider only a single branching (cf. \ Fig.\ \ref{fig:P1}). This single branch is denoted as a scenario consisting of the $N$ nodes from the root node to the leaf node. Each scenario $s \in \mathcal{S} $ has its own set of states $\vec{z}_{k,s} \in \mathbb{R}^{n_z}$ (i.e. the respective nodes) and control inputs $\vec{u}_{k,s}\in \mathbb{R}^{n_u}$, where $n_z, n_u, |S|$ are the number of states, control inputs and scenarios, respectively.  We defer our discussion of which subset $\mathcal{S}$ of modes from $\mathcal{M}$ to include in the scenario tree to sec. \ref{sec:scen}.
We define the set $I$ to encapsulate all elements $k,s$ within the scenario tree. This gives us a discrete-time formulation of the dynamic system using eq. (\ref{eq:2}) 
:
\begin{equation}
\vec{z}_{k+1,s}= f(\vec{z}_{k,s},\vec{u}_{k,s})   
\end{equation}
In formulating the optimization problem, we employ the running cost and constraint from the nominal MPCC as defined in sec. \ref{sec:nominal}.
\begin{subequations}
\label{eq:NLP}
\begin{align}
 &\underset{\vec{z}_{k,s},\vec{u}_{k,s} \forall (k,s) \in I}{\min} \  \sum_{s \in\mathcal{S }}\omega_s \sum_{k=0}^{N-1} J_k(\vec{z}_{k,s}, \vec{u}_{k,s})  \\
 & s.t.  \   \vec{z}_{k+1,s}= f(\vec{z}_{k,s},\vec{u}_{k,s}) \ \forall (k,s) \in I  \\
 & \vec{z}_{0,s} = \vec{z}(0) \ \forall s \in \mathcal{S} \\
 & \vec{z}_{k,s} \in  \mathcal{Z}(\vec{o}_{k,s}), \ \vec{u}_{k,s} \in \mathcal{U} \ \forall (k,s) \in I \\
 & \vec{u}_{k,i} = \vec{u}_{k,j} \ \forall (i,j)\in \mathcal{S} \  \forall \ k \in [0, b]
\end{align}
\end{subequations}
where $\mathcal{Z}$ is a set of state constraints, including the obstacle constraints for each scenario, and $\mathcal{U}$ is the set of control input constraints.
The process for selecting scenarios and defining the cost weight $\omega_s$ is detailed in Section \ref{sec:scen}.

Each branch's states and control inputs adapt to the selected prediction (mode) of the obstacles $o$ to satisfy the constraints and optimize the cost of the respective scenario, only.
However, the BMPCC includes the constraints in eq. (\ref{eq:NLP}e) which are called non-anticipatory 
\begin{remark}The non-anticipatory constraints force the control inputs of all scenarios to equal each other in the first timesteps $[t_0, t_b]$, i.e., the so-called branching time. These are introduced to model that the controller cannot \emph{anticipate} in the beginning which of the scenarios will occur.
\end{remark}

This setup, termed passive information gathering \cite{tas_decision-theoretic_2023}, models the intuition that more information on scene evolution will be available in future. Up to time $t_b$, the AV follows a consensus plan and postpones commitment to a single mode. After the branching time $t_b$ these constraints in eq.\ (\ref{eq:NLP}e) are no longer active. We detail this postponing in sec. \ref{sec:adp}.

\section{Scenario Selection}\label{sec:scen}
This section describes the construction of the scenario tree. To achieve this, all $|\mathcal{M}|$ predictions outputted by the neural network in sec. \ref{sec:predictor} are considered. The objective is to limit the number of considered scenarios  $\mathcal{S}$ by a small subset of the predicted modes,  $\mathcal{S} \subseteq \mathcal{M}$, $|\mathcal{S}| \leq |\mathcal{M}|$, to make the BMPCC tractable for real-time applications while still not excluding scenarios that may restrict the robustness of the controller.
For that, initially, the predicted trajectories that have similar effects on the planner are clustered in sec. \ref{sec:uvd}. Then, the most risk-relevant prediction from each of these clusters is selected and added as a scenario in the scenario tree in sec. \ref{sec:risk}.
To determine these similarities and relevance of predictions for the AV, i.e., for both of our methods presented subsequently, the future behavior of the ego vehicle needs to be known. Since the current motion plan $\vec{\tau}^*_t$ is not available in advance, we make the following assumption:

\begin{assumption}The optimal motion plan   calculated in the last planning cycle $\vec{\tau}^*_{t-1}$ is sufficiently
close to the current optimal motion plan $\vec{\tau}^*_{t}$ to judge the topological relationships of the agents.
\end{assumption}

\begin{figure}[t]
\centering
    \includegraphics[trim = 0mm 0mm 0mm 0mm, clip, width=0.45\textwidth]{"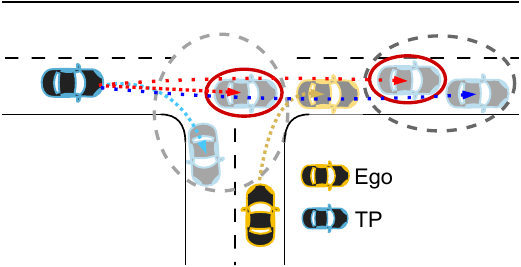"}  
    \caption{Illustration of the Scenario Selection procedure. Predictions enclosed in grey ellipses belong to same cluster. Red ellipses mark scenarios selected from each cluster.}
    \label{fig:P2} 
    \vspace{-1.3em} 
\end{figure}

\subsection{Topology-based Clustering} \label{sec:uvd}
For clustering the predictions, we examine the topology of individual predicted trajectories. The underlying idea is based on the insight that certain obstacle behaviors, even if not identical, lead to similar optimal AV behavior, while others require fundamentally different responses. An example of that is provided in Fig.\ \ref{fig:P2} where the ego vehicle must either brake or accelerate at the intersection to maintain a safe distance from the TP depending on the TP´s behavior\footnote{In this example, stopping avoids collision in both cases but is far from being optimal due to excess of conservatism.}. Our methodology relies on grouping predictions of TPs according to this criterion, considering exactly one scenario per cluster $\mathcal{C}_s$ where $\bigcup_{s \in \mathcal{S}} \mathcal{C}_s = \mathcal{M}$ i.e. leading to a partition of the set $\mathcal{M}$. To find these clusters, we employ uniform visibility deformation (UVD) to assess the similarity of predicted trajectories.
\begin{definition}
Two trajectories $\vec{\tau}_1(\alpha)$,$\vec{\tau}_2(\alpha)$ parameterized by $\alpha \in [0,1]$ and satisfying $\vec{\tau}_1(0)=\vec{\tau}_2(0)$, $\vec{\tau}_1(1)=\vec{\tau}_2(1)$ belong to the same \emph{uniform visibility deformation}, if for all $\alpha$, the line connecting $\vec{\tau}_1(\alpha)$,$\vec{\tau}_2(\alpha)$ is collision-free.\cite{singh}
\end{definition}

UVD is typically utilized to compare ego trajectories for global trajectory planning, aiming to efficiently identify multiple topologically distinct trajectories\cite{singh, deGroot}. We use this method not to compare ego trajectories but to cluster TP trajectories (i.e., the ego serve as obstacle). For this purpose, the final position condition ($\vec{\tau}_1(1)=\vec{\tau}_2(1)$) is irrelevant to us, so we disregard it.

The precise procedure for clustering entails checking whether two predicted trajectories belong to the same group by uniformly discretizing them and selecting equidistant timesteps as the parametrization. Subsequently, we connect corresponding points on the two trajectories, i.e., points at the same timestep, and determine if these lines intersect with the ego vehicle motion plan $\vec{\tau_{t-1}}$ (s.\ Fig.\ \ref{fig:method}). Thus, they do not belong to the same cluster if at least one connected line intersects with the ego. In the case of multiple TPs, if it is found that two trajectories are not topologically equivalent for at least one TP, the respective predictions of all TPs are assigned to different clusters.

\subsection{Ranking and Selection based on Collision Risk}\label{sec:risk}

The aim of building the scenario tree is to map various reactions of the AV that cover possible behaviors of the TPs for safe and comfortable driving. Therefore, we add one prediction from each of the formed clusters to the scenario tree. The most relevant prediction that best represents the entire cluster must be selected. Representation here means that if the selected scenario is taken into account in the planning, i.e. to avoid collisions, the collision with the other predictions in the respective cluster are implicitly also likely avoided (s.\ Fig.\ \ref{fig:P2}). For this process, we introduce the criterion of collision risk. We utilize the work of \cite{philipp_analytic_2019}, who presents an efficient way to calculate the collision event probability between two oriented rectangles analytically.
\begin{definition}
The probability density of a collision event between a TP $\vec{o_{k,m}}$ and the ego $\vec{z_{k}}$ at a certain timestep $t_k$ is called Collision Event Probability (CEP) density $\frac{dC^o_m(t_k)}{dt}$. The CEP $C^o_m(t_i,t_j)$, the probability of a collision during a period of time $[t_i, t_j]$, i.e., between the ego plan and TP prediction, is the integral of the CEP density over time.
\end{definition}
Leveraging the procedure detailed in \cite{philipp_analytic_2019}, we estimate $\frac{dC^o_m(t_k)}{dt}$ based on a first-order Taylor approximation. The formula depends on the TP's pose, velocity, and the respective Gaussian uncertainty at timestep $t_k$. This we can extract from the output of our prediction in sec. \ref{sec:predictor}.
Furthermore, it depends on the ego plan, i.e., the pose and velocity of the AV at timestep $t_k$. The trajectory of the ego vehicle is considered deterministic where we rely on Assumption 1. 

To obtain the CEP, we accumulate the timesteps over the prediction horizon with the sample time $\Delta t$:
\begin{equation}
   C^o_m(t_0,t_{N-1}) = \sum_{k=0}^N  \frac{dC^o_m(t_k)}{dt} \Delta t 
\end{equation}

We calculate this collision risk for each prediction of each TP. This collision risk is utilized to define a metric for ranking and selecting the predictions. An exemplary decision function for this purpose may be defined as follows:
\begin{equation}\label{eq:D}
   D^o_m =  C^o_m(t_0,t_N) + \lambda \pi^o_m
\end{equation}
The choice of the decision function is motivated to avoid excluding very likely predictions (i.e. high occurrence probability $\pi^o_m$) that have a low collision probability with the AV. If one would only consider trajectories with a high collision risk and exclude the very likely trajectory, one would adopt an excessively conservative approach. Accordingly, the tuning factor $\lambda$ may be adapted as trade-off between conservatism and effectiveness.
Based on this decision metric, for each cluster $\mathcal{C}_s$, the mode  with the maximum value is introduced as scenario $s$ into the scenario tree.
\begin{equation}
s = \underset{m \in \mathcal{C}_s}{\text{argmax}}  \sum_{o=0}^{O-1} D_{m}^{o}
\end{equation}

Once the scenario tree is constructed, the weighting $\omega_s$ for each scenario in the cost function in eq. (\ref{eq:NLP}a) can be determined. This weighting is derived through the sum of the probabilities of the individual predictions per cluster.

\section{Adaptive Decision Postponing}\label{sec:adp}
This section outlines how the branching time, introduced in Remark 1, is determined within our framework.
The underlying assumption of the BMPCC is that after the branching time is reached the uncertainty (which of the predictions is the correct one) resolves, and the AV knows which option in the scenario tree it should take. This uncertainty depends on various factors of the traffic scene and cannot be predetermined offline independently of the current situation. Thus, this necessitates to adapt the branching time online based on the observed situation and depending on when it is expected to have sufficient information for a decision. A wrong selection of the branching time can harm the AV performance. If set too short, decisions may be made prematurely, potentially resulting in the selection of the wrong scenario due to the remaining uncertainty. 
A longer branching time postpones the decision, aiming to accommodate all scenarios, but this can result in overly conservative behavior where an opportunity may be missed, e.g., to merge into a promising gap.
 
\begin{figure}[t]
\centering
    \includegraphics[trim = 0mm 0mm 0mm 0mm, clip, width=0.45\textwidth]{"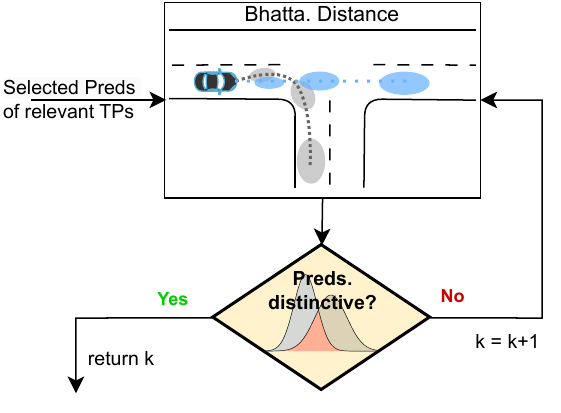"}
    \caption{Procedure of the online estimation of the branching time based on the Bhattacharya Distance}
    \label{fig:P3} 
    \vspace{-1.3em} 
\end{figure}

To determine the branching time, we estimate after how many timesteps the AV can confidently assign the TPs state measurement to one of the predicted trajectories. For this purpose, we introduce the Bhattacharyya distance \cite{bhatta}.
\begin{definition}
The Bhattacharyya Distance $B$ is a measure to quantify the similarity between two probability distributions $p(x)$ and $q(x)$,   $B = -\ln(\int(\sqrt{p(x)q(x)}dx))$.
\end{definition}
We leverage this measure to assess for every timestep in the prediction horizon how much two Gaussian distributions  $\mathcal{N}^o_{k,i}, \mathcal{N}^o_{k,j}, \ i,j \in \mathcal{S}$ from the GMM in eq. (\ref{eq:gmm}) overlap (s.\ Fig.\ \ref{fig:P3}).
For two multivariate Gaussian distributions, the formula simplifies to:
\begin{equation}
    \begin{split}
    B(o,k,i,j) &= \frac{1}{8} \left(\boldsymbol{\mu}^o_{k,i}-\boldsymbol{\mu}^o_{k,j}\right)^T \left(\boldsymbol{\Sigma}^o_k\right)^{-1}\left(\boldsymbol{\mu}^o_{k,i}-\boldsymbol{\mu}^o_{k,j}\right) \\
    &\quad +\frac{1}{2} \ln \left(\frac{\operatorname{det} \boldsymbol{\Sigma}^o_k}{\sqrt{\operatorname{det} \boldsymbol{\Sigma}^o_{k,i} \operatorname{det} \boldsymbol{\Sigma}^o_{k,j}}}\right)
    \end{split}
\end{equation}
where $\boldsymbol{\Sigma}^o_k =\frac{\boldsymbol{\Sigma}^o_{k,i} + \boldsymbol{\Sigma}^o_{k,j}}{2}$. If a predefined threshold is exceeded $B(o,k,i,j) \geq B_{th}$ for a certain timestep we set the branching time at that particular time $t_b$. If more than one TP is in the scene, we use the previously calculated collision risk to threshold initially which  TPs are relevant for decision postponing. After that, if there are more than two TPs left, we calculate the time $t^o_b$ where $B(o,k,i,j) \geq B_{th}$ for each of these TPs and simply select the maximum branching time $t_b =\text{max}_o t_b^o$. Similarly, if  $|\mathcal{S}|>2$, we calculate the distinguishability for all combinations and select the maximum branching time.

\section{Performance Evaluation} \label{sec:Results}
Our evaluation is divided into qualitative and quantitative results. Qualitatively, we analyze our methods in a traffic intersection. For quantitative results, we use Monte Carlo analysis of random highway merging scenarios. We implement our framework in CasADi using the IPOPT solver for the BMPCC. We use a prediction horizon of $N=40$ and a sample time of $\Delta t = 100  \ ms$ for both our framework and the baselines. We limit the number of scenarios selected by our method to $S_{max} = 2$ and show that it can perform better than using BMPCC with up to 4 scenarios. If our approach outputs more than $S_{max}$ clusters $\mathcal{C}_s$, we select the clusters that have the highest relevance based on our metric in eq.(\ref{eq:D}). 

Our first Baseline is the conventional MPCC (\emph{CMPCC}) introduced in sec. \ref{sec:nominal}. 
Second, we use Scenario-based MPCC (\emph{SCMPCC}), which takes the 5  TP predictions with highest probability $\pi^m_o$, i.e., only excluding very unlikely predictions. The main difference to the BMPCC is that SCMPCC does not consider future feedback. In other words, the SCMPCC equals the BMPCC only if the branching time is set identical to the prediction horizon ($t_b=t_N$).
The following variants are BMPCCs where we ablate different parts of our method.
\emph{BMPCCnoSS2}, \emph{BMPCCnoSS3} and \emph{BMPCCnoSS4} leaves out the scenario selection and decision postponing part and only add the $S_{max}=2$, $S_{max}=3$, and $S_{max}=4$ scenarios with highest probability to the scenario tree. 
Then, in \textit{BMPCCCnoDP}, we employ the scenario selection with $S_{max}=2$ but leave out the adaptive decision postponing. For all four of these variants, the branching time is set to the very first timestep as done in  \cite{fors_resilient_2022, chen_interactive_2021,alsterda_contingency_2019}.
\begin{table*}[t]
    \vspace{0.5em} 
    \caption{Comparison using Monte Carlo analysis of Highway Merging Scenarios}
    \vspace{-0.6em} 
    \centering
    \begin{tabular}{ |p{5.5cm}||m{1.7cm}|m{1.7cm}|m{1.7cm}|m{1.7cm}|m{2.8cm}|  }
 \hline
  & \multicolumn{3}{c|}{\textbf{Merging Execution}} & \multicolumn{2}{c|}{\textbf{Convergence Quality}} \\ 
 \cline{2-6}
  & Success & Aborted & Collision &  Mean Cost &  Mean Solving time (ms) \\ 
 \hline
 \textbf{CMPCC (1 Scen.)} & 89\% & \textbf{5\%} & 6\% &  1397.6  & \textbf{22} \\ 
  \hline
   \textbf{SCMPCC} & 86\% & 14\% & \textbf{0}\% & 1226.5   & 73 \\
  \hline
\textbf{BMPCC no Scenario Selection (2 Scen.) } & 89 \% & 8 \% & 3 \% & 1039.9   & 50 \\ 
  \hline
 \textbf{BMPCC no Scenario Selection (3 Scen.)} & 89 \% & 9 \% & 2 \% & 883.4  & 92 \\ 
  \hline
   \textbf{BMPCC no Scenario Selection (4 Scen.)} & 90 \% & 9 \% & 1 \% &  1074.0 & 167 \\ 
  \hline
   \textbf{BMPCC no adapt. Dec. Postponing (2 Scen.)} & 89 \% & 9 \% & 2 \% &  817.8 & 64 + 9 (BMPCC+ rest) \\ 
  \hline
 \textbf{Our Framework (2 Scen.)} & \textbf{94 \%} & \textbf{5\%}   & 1\% &  \textbf{575.5}  & 54 + 10 (BMPCC+ rest)\\   
 \hline
\end{tabular}
\label{tab:mca}
\vspace{-1.0em} 
\end{table*}
\subsection{Traffic Intersection} 
\begin{figure}[]
\centering
    \includegraphics[trim = 0mm 0mm 0mm 0mm, clip, width=0.48\textwidth]{"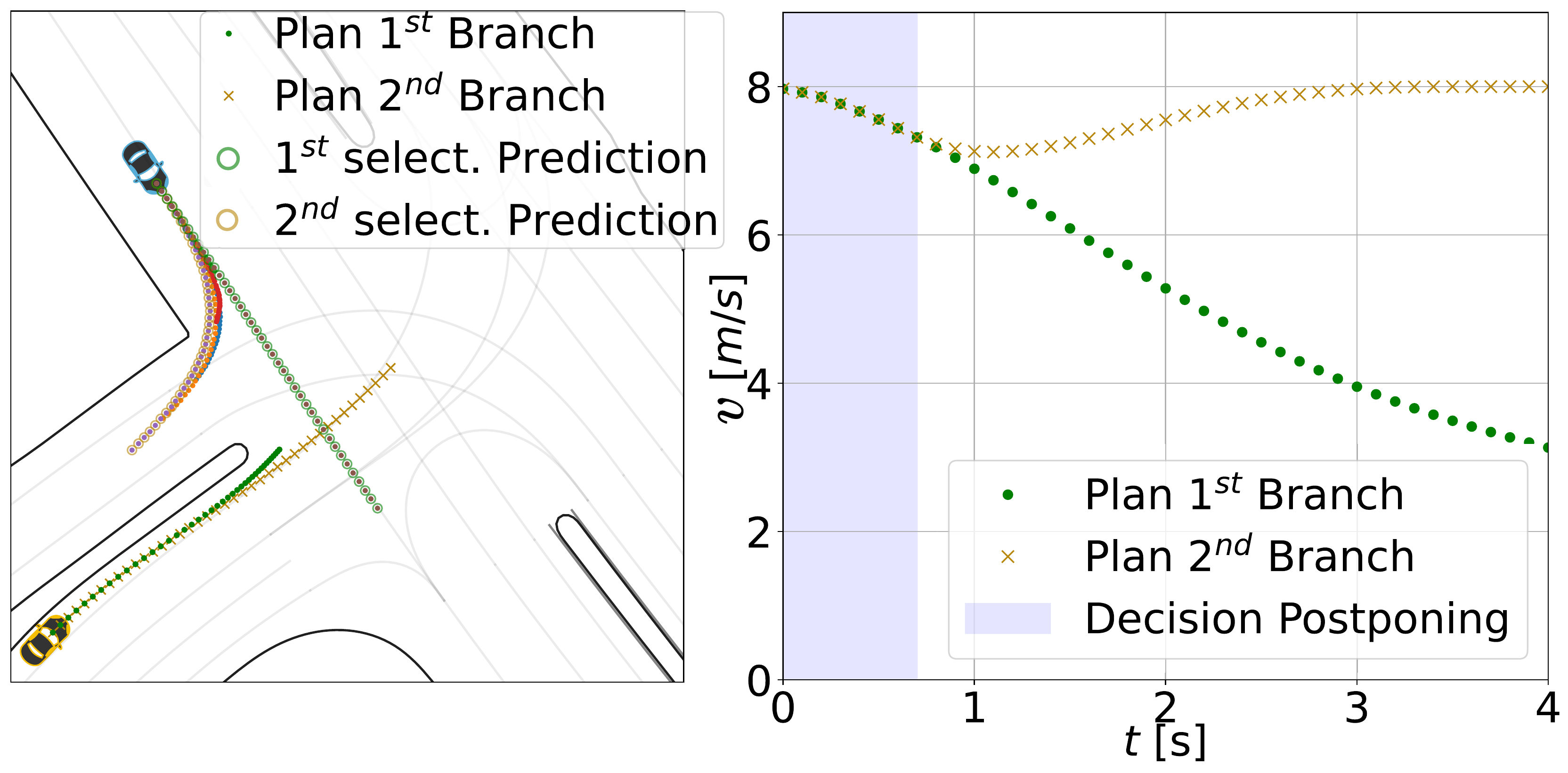"}
    \vspace{-1.3em} 
    \caption{Prediction of TP (blue car)  and the plan of the AV (yellow car) at a certain timestep. Predictor outputs 6 predictions (shown in different colors) from which 2 most-relevant are selected. Corresponding  trajectories are shown in corresponding colors. \textbf{Right:} Planned velocity profile.}
    \label{fig:P4} 
    \vspace{-1.3em} 
\end{figure}
\begin{figure}[h]
\centering
\vspace{-0.4em} 
    \includegraphics[trim = 2mm 0mm 0mm 0mm, clip, width=0.49\textwidth]{"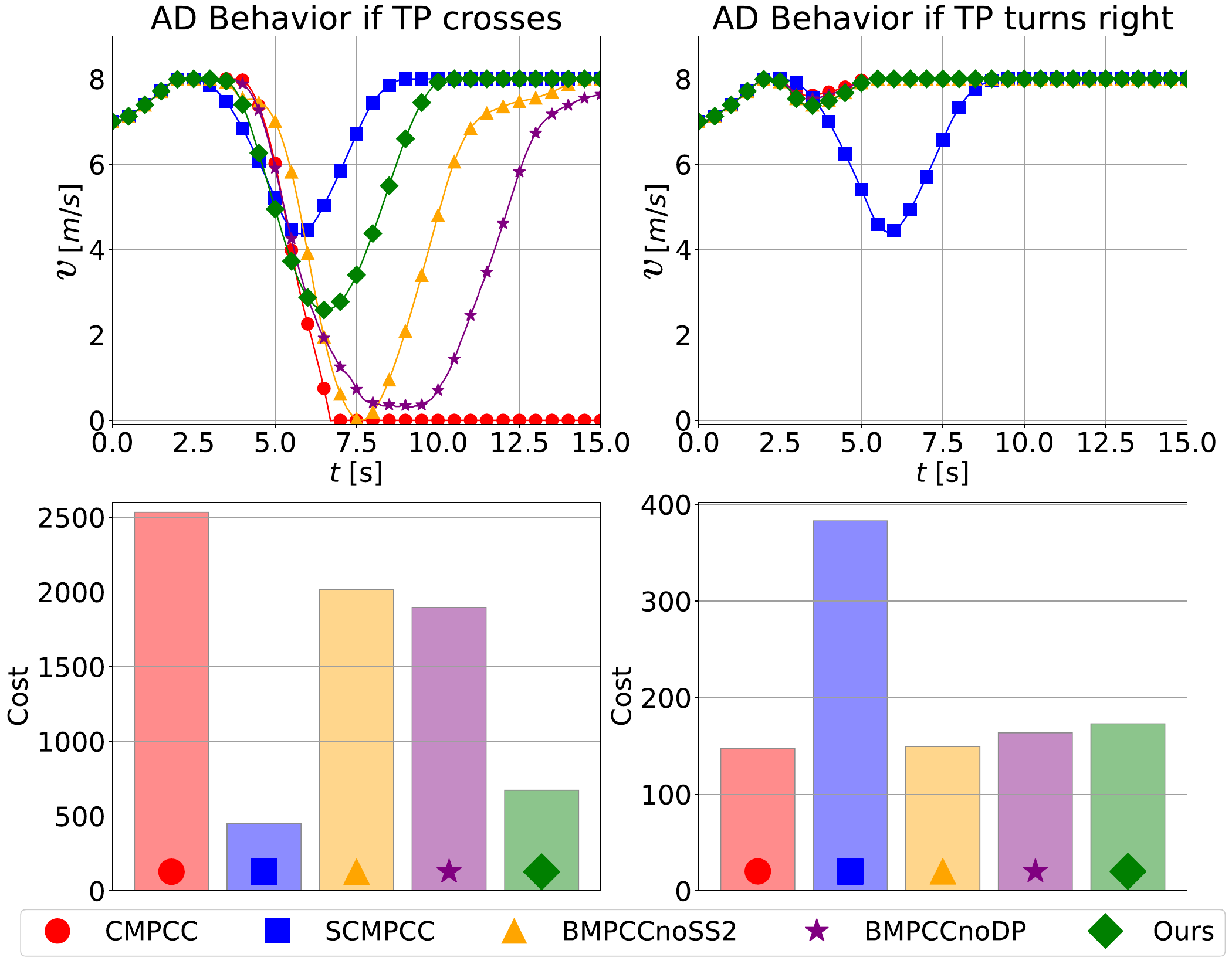"}
    \vspace{-1.5em} 
    \caption{Qualitative comparison of the performance of our approach and the Baselines in the traffic intersection scene }
    \label{fig:P5} 
    \vspace{-0.83em} 
\end{figure}

In this traffic scenario, both our AV and a TP approach an intersection where the TP has the right of way. Our predictor predicts multiple trajectories for the TP that either turns right before the intersection or crosses it (s.\ Fig.\ \ref{fig:P4}). Accordingly, as depicted in  Fig.\ \ref{fig:P4}, our approach estimates that uncertainty resolves after $t_b=0.8s$ and only slightly brakes to accommodate both possible TP behaviors. We run the scenario twice, once where the TP turns right before the intersection and once where it crosses. In Fig.\ \ref{fig:P5}, the performance of our method is compared to the other methods. Costs are calculated for the entire scenario duration, following the same criteria (comfort, reference tracking, etc.) as defined in eq. (\ref{eqn:J}). When the TP crosses the intersection, CMPCC and BMPCCnoSS2 perform poorly because the most probable predictions indicate a right turn in the beginning. In the case of CMPCC, emergency braking is even required to avoid a collision. Even when the most-relevant scenarios are selected with our method, BMPCCnoDP performs poorly because it always expects uncertainty to resolve in the next time step, which is not the case. As a result, it starts braking later than our framework, leading to significantly less comfortable behavior. SCMPCC shows the best behavior for this scenario because it plans for the entire prediction horizon for the crossing scenario. However, in the scenario where the TP turns right, SCMPCC exhibits the worst behavior because it plans for the worst-case scenario (TP crosses intersection) for a long time, i.e., too conservatively. The other methods all exhibit satisfactory behavior for this scene. In conclusion, our method succeeds in both cases, while the others only perform well in one of them.

\subsection{Highway Merging} 
We conduct a Monte Carlo analysis, sampling 100 random highway merging scenarios (s. Tab. \ref{tab:mca}). Fig.\ \ref{fig:P5} depicts an example of the evolution of a single merging scene using our method. To simulate TP behavior, we utilize the Intelligent Driver Model(IDM), with IDM parameters randomly sampled. For method comparison, we assess the rate of merging successes, occurrences of getting stuck on the merging lane (aborted), and collisions. Additionally, we evaluate computation time (including time for scenario selection and adaptive decision postponing) and, again, the costs. Results indicate, as expected, that SCMPC is very safe but tends to be overly conservative. Our approach achieves a balance between robustness and performance.
The impact of scenario selection is best observed in the comparison between BMPCCnoDP and BMPCCnoSS (with 2, 3, and 4 scenarios). With scenario selection, BMPCCnoDP partly outperforms BMPCC with 3 or 4 scenarios, with significantly reducing computation time. Furthermore, the distinct effect of adaptive decision postponing is evident in the comparison between our framework and BMPCCnoDP.

\begin{figure}[]
\centering
    \includegraphics[trim = 2mm 0mm 0mm 0mm, clip, width=0.45\textwidth]{"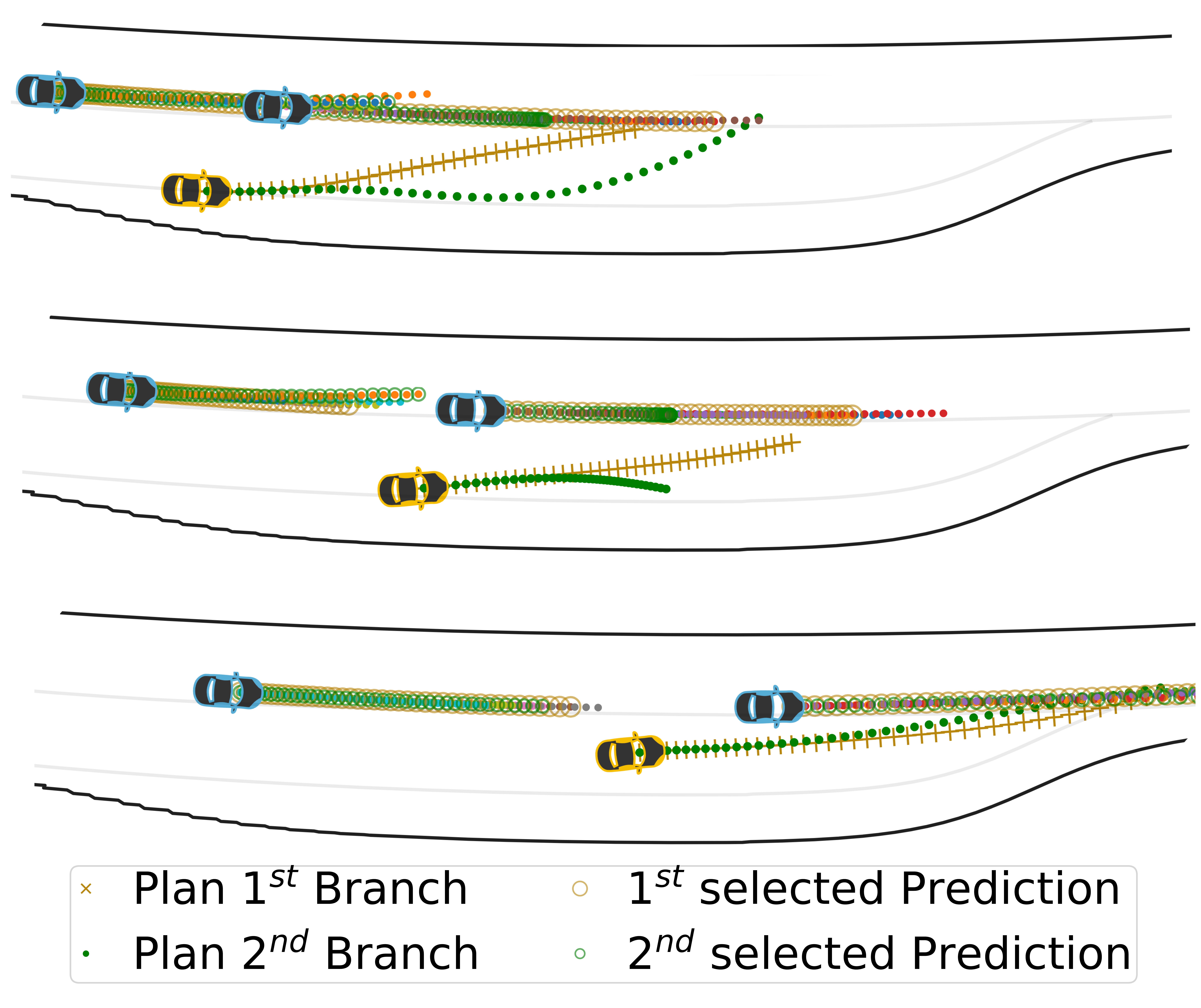"}
    \vspace{-0.5em}
    \caption{ AV behavior at three different timesteps. Initially, the AV  does not commit to merging in front or behind the first TP. As uncertainty reduces, the branches converge.}
    \label{fig:P6} 
    \vspace{-1.5em} 
\end{figure}
\section{Conclusion}\label{sec:conclusion}
We proposed a framework for motion planning and control that addresses multi-modal uncertainty of traffic participants.  By integrating a learning-based multi-modal motion predictor into a Branch MPCC we improve AV safety without excessive conservatism. However, this enhancement required designing intermediary modules. 
Specifically, we developed a scenario selection method leveraging topology and risk metrics as selection criteria of multi-modal predictions. By reducing the number of considered predictions, we rendered Branch MPC tractable for real-time applications. Our findings demonstrate that few predictions are sufficient if the selected predictions cover this uncertainty well enough. Furthermore, we introduced an adaptive decision-postponing approach to estimate when sufficient future information allows to commit to one of the predictions, i.e., when branching in the Branch MPCC should begin. Our evaluation has shown that this also has a major impact on the performance.

\bibliographystyle{IEEEtran}
\bibliography{Refs}
 
\end{document}